\documentclass[letterpaper,10pt,conference]{ieeeconf}
\IEEEoverridecommandlockouts
\usepackage[T1]{fontenc}
\usepackage{amsmath,amssymb,graphicx,array,tabularx,booktabs}
\newcolumntype{L}[1]{>{\raggedright\arraybackslash}p{#1}}
\usepackage{cite}
\usepackage{balance}
\usepackage[hidelinks]{hyperref}
\newcolumntype{Y}{>{\raggedright\arraybackslash}X}
\hypersetup{pdfauthor={Ben Ma and Jixuan Cai},pdftitle={RoboJackVR: Turning Robot Tasks into VR Games for Scalable Demonstration Collection},pdfsubject={Preprint},pdfkeywords={Human-to-Robot Learning, Virtual Reality, Demonstration Collection, Gamification, Imitation Learning}}
\title{\LARGE\bf RoboJackVR: Turning Robot Tasks into VR Games\\for Scalable Demonstration Collection}
\author{Ben Ma and Jixuan Cai\\[3pt]
CosmicFlow Technology\\[3pt]
{\small\texttt{qwsniuab@gmail.com}, \texttt{codyjcai@gmail.com}}}
\begin{document}
\maketitle
\thispagestyle{plain}
\pagestyle{plain}
\bstctlcite{ral-bib-control}

\begin{abstract}
Scaling robot demonstration datasets requires collection systems that
accommodate more participants and diverse tasks, yet teleoperation barriers
and task-adaptation effort limit this expansion. We present RoboJackVR, a
standalone virtual reality (VR) game in which players manipulate simulated
robots to execute combat skills and contribute demonstrations. MuJoCo
simulation and demonstration recording run locally on the headset, removing
the need for an external PC or simulation server during collection. Tutorials, skill
selection, combat consequences, and character progression organize repeated
task attempts within a continuing game. Task contracts declare assets,
controls, resets, success criteria, and recording semantics; separate game
configurations specify guidance, attempt groups, skill effects, and course
placement. Shared bindings connect these declarations to task execution,
outcome handling, and recording, allowing tasks to reuse the game flow while
retaining their control and success semantics. Batch tools package supported
resources and generate configurations, with manual adaptation and review for
new interfaces and unsupported predicates. We instantiate the approach with
SO101-Nexus-derived and LIBERO tasks. On the 66 LIBERO-90 tasks covered by the collected data, ACT policies
trained solely on VR demonstrations retain approximately 91\% of the
full-standard simulation-demonstration baseline's success rate on held-out
initial states, averaged over four training seeds. These results support the
utility of play-collected VR demonstrations for policy learning in this setting.
\end{abstract}
\smallskip

\section{Introduction}
\label{sec:introduction}

Scaling robot demonstration collection requires broader participation and
more task coverage, but teleoperation barriers and the effort of adapting
new tasks limit both~\cite{robocrowd2024,iris2025}. DROID illustrates the
value of diverse demonstrations alongside the hardware, labor, and
coordination needed to obtain them~\cite{droid2024}. Extended reality (XR) interfaces to
simulation may still depend on an external simulation host~\cite{iris2025};
running task execution on the headset removes this additional computing
requirement from the participant's collection setup. Accessible interfaces
and reusable task infrastructure address parts of this challenge; games
provide a way to organize repeated contributions within an activity that
players can understand and choose to continue.

We ask: \emph{How can existing robot task collections become VR game content
through a reusable gameplay and collection workflow, and can demonstrations
collected during play support robot learning?} Our central idea gives each
task two connected roles: an operation challenge that players choose and
master, and a demonstration unit with explicit controls, success criteria,
and recording semantics. The design must connect task execution to game
consequences while preserving the information needed to interpret the data.

\begin{figure*}[t]
  \centering
  \includegraphics[width=\linewidth]{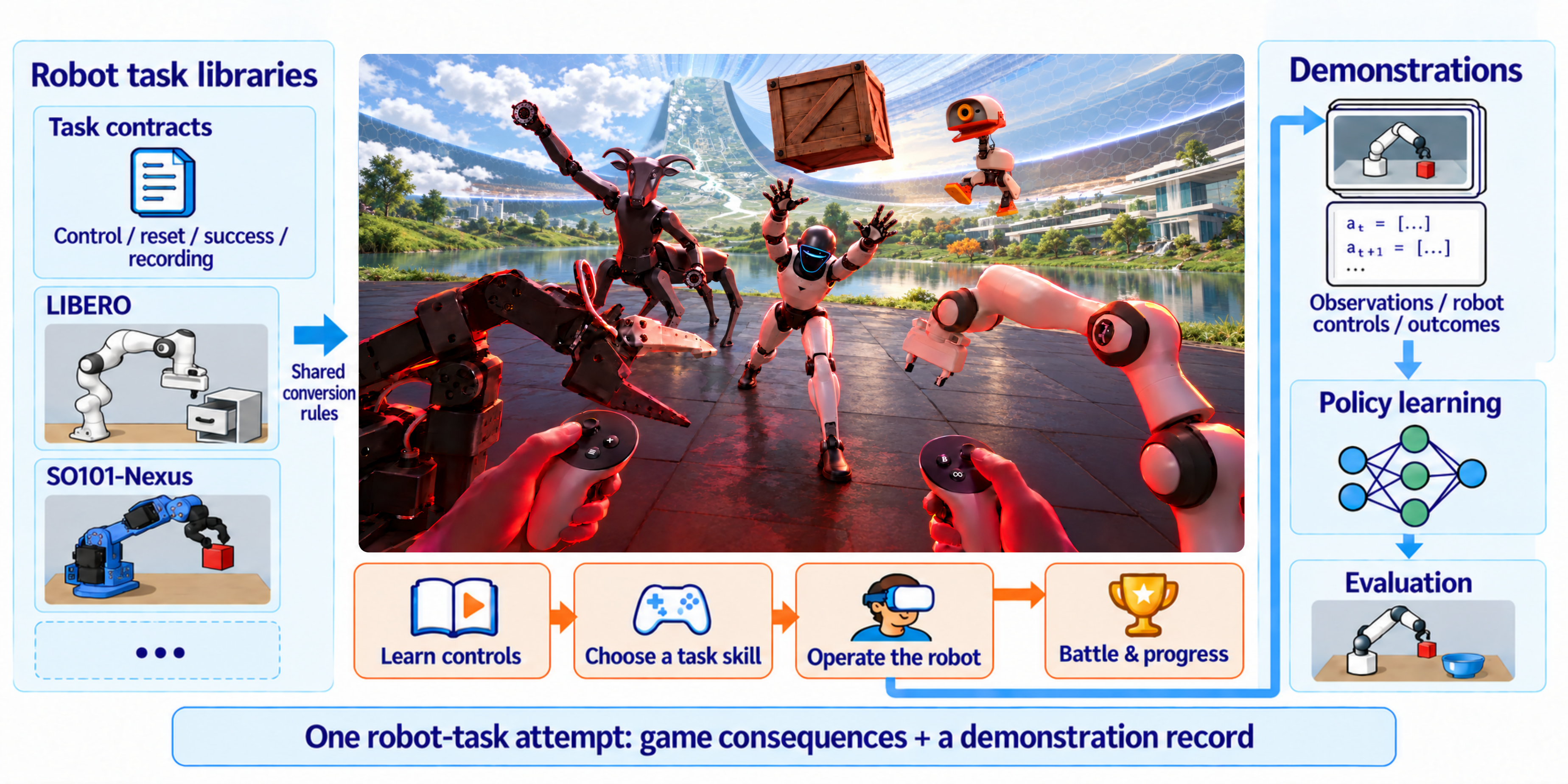}
  \caption{Overview of RoboJackVR. Robot task libraries enter a shared VR game
  through task contracts and conversion rules. Players learn controls, choose
  task skills, and operate robots; a single task attempt both produces game
  consequences and records a demonstration for downstream learning and
  evaluation. The headset executes the game and robot simulation and records
  raw demonstrations; a separate host processes data and trains policies.
  The central scene is a conceptual illustration; task and learning icons are
  schematic.}
  \label{fig:framework}
\end{figure*}

We present RoboJackVR, a VR adventure game built around robot manipulation.
Players select skills according to the battle state, operate simulated robots,
and progress through subsequent challenges. Opening a drawer, for example,
executes an attack skill: the task outcome changes the battle state and
produces a demonstration record. Task contracts declare execution and
recording interfaces; separate game configurations specify guidance, attempt
groups, skill effects, and course placement. Shared bindings connect them to
instruction, control, feedback, progression, and recording
(Figure~\ref{fig:framework}). New tasks can extend the same game while
retaining their task-specific control and success semantics.

The paper focuses on expanding playable task coverage, instantiated with
SO101-Nexus-derived~\cite{so101nexus} and LIBERO tasks~\cite{libero2023}. Concrete bindings
identify reusable components and the configuration or implementation that
remains task-specific. A policy-learning study on LIBERO-90 using Action
Chunking with Transformers (ACT)~\cite{zhao2023act} compares full and reduced
standard-data budgets, VR-only training, and joint training on VR and reduced
standard data. Standard demonstrations here are LIBERO's public
\emph{simulation} data, not physical-robot recordings. The study connects
on-headset demonstration collection, trajectory conversion, and downstream
policy training within a common experimental pipeline.

Our contributions are:
\begin{itemize}
  \item A reusable task-to-game interface that preserves declared task controls
  and success conditions while connecting robot-task execution to shared
  guidance, combat feedback, progression, and demonstration recording.
  \item A standalone VR game that runs MuJoCo simulation and demonstration
  recording on consumer Quest headsets, with default combat configurations
  for 130 LIBERO and four SO101 tasks.
  \item A pipeline that converts VR records into the standard LIBERO
  demonstration format, and a LIBERO-90 ACT study comparing
  standard-data budgets, VR-only training, and joint VR and standard-data
  training.
\end{itemize}

\section{Related Work}
\label{sec:related}

\paragraph{Human participation in demonstration collection.}
\label{sec:related-participation}
DROID distributes collection across institutions and physical settings,
whereas UMI uses hand-held grippers without a robot at the collection
site~\cite{droid2024,umi2024}. RoboTurk combines smartphone teleoperation
with cloud infrastructure, and early VR work identified gaming communities
as potential contributors~\cite{roboturk2018,crowdvr2018}. DART combines
augmented reality (AR) teleoperation and cloud simulation; COBALT adds concurrent collection
and contributor training~\cite{dart2024,cobalt2026}. RoboCrowd studies
public participation and incentives, including crowd-data pretraining
followed by expert-data fine-tuning~\cite{robocrowd2024}. These approaches
motivate connecting access, contributor experience, and downstream data use.
RoboJackVR organizes repeated task attempts within a continuing VR game.

\paragraph{Games and play for robot learning.}
\label{sec:related-games}
Games with a Purpose and gamification research connect useful activity to
game design, while motivation research emphasizes autonomy and
competence~\cite{gwap2008,gamification2011,motivation2006}. In robot
learning, Play-LMP learns from unsegmented teleoperated play, and MimicPlay
combines human-hand play videos with limited robot
demonstrations~\cite{playlmp2020,mimicplay2023}. Our focus is explicit
game rules for interaction with existing robot tasks.

The closest precedents connect game content to downstream learning.
RoboCade embeds games in remote real-robot teleoperation and co-trains on
support tasks sharing manipulation skills with target tasks~\cite{robocade2025}.
Project Kitchen collects bare-hand VR cooking interactions, extracts contact
points and sub-goals, and transfers affordance supervision to policies
trained with limited robot data~\cite{projectkitchen2026}. Other VR robot
games study motion and workspace coverage, while CrowdHRI uses multiplayer
mixed reality to collect interaction data~\cite{vrrobotgames2026,crowdhri2025}.
RoboJackVR uses existing task collections as content for a complete adventure
game: task operations execute selectable combat skills, and outcomes connect
to progression and later choices. The reusable unit is a task attempt with
declared execution, success, and recording semantics.

\paragraph{Reusable task and interaction infrastructure.}
\label{sec:related-reuse}
robosuite, RLBench, and LIBERO provide simulation components, task definitions,
and demonstration benchmarks~\cite{robosuite2020,rlbench2019,libero2023}.
RoboVerse's MetaSim unifies task and simulation APIs; IRIS and XRoboToolkit
support reusable immersive interaction~\cite{roboverse2025,iris2025,xrobotoolkit2025}.
MimicGen expands demonstrations across scene configurations, while AXIS
combines community collection with automated task generation and
validation~\cite{mimicgen2023,axis2026}. RoboJackVR addresses the connection
between task infrastructure and a shared game: adapters retain task-specific
execution, while bindings reuse instruction, feedback, progression, and
recording. New interfaces and unsupported task semantics still require adaptation.

\section{RoboJackVR}
\label{sec:proposal}

RoboJackVR gives each robot-task attempt two linked roles: a game action that
changes combat or progression, and an episode that records robot behavior.
Task contracts provide the integration boundary; instruction, battle rules,
and progression turn task execution into play, and recording retains its
learning interface.

\paragraph{On-headset execution.}
RoboJackVR links Unity to Android/ARM64 native libraries for robot control,
MuJoCo dynamics~\cite{todorov2012mujoco}, and task evaluation. Unity handles XR input and game
presentation, while raw demonstrations are saved on the headset. Once the
app, task resources, and user identity are set up, local collection
requires no external PC or remote simulation server. Account services and
uploads use network connections; image generation, action conversion, and
policy learning run off-device.

\begin{table*}[t]
  \caption{Two task-integration examples showing task-specific settings,
  shared bindings, and required checks (catalog v1.3.1).}
  \label{tab:integration-examples}
  \centering\small
  \setlength{\tabcolsep}{4pt}
  \renewcommand{\arraystretch}{1.12}
  \begin{tabularx}{\linewidth}{@{}L{0.19\linewidth}YY@{}}
    \hline
    \textbf{Binding} & \textbf{SO101: grasp and lift} & \textbf{Franka: middle drawer} \\
    \hline
    Task contract &
    Arm/cube assets; reset and grasp/lift requirements. &
    Cabinet assets; source initial-state bank; drawer-opening requirement. \\[2pt]
    Task adapter &
    SO101 control; reset and grasp/lift evaluation. &
    Franka Direct control; initial-state restoration and drawer-opening evaluation. \\[2pt]
    Control/record &
    Versioned SO101 profile; six-component state/command schema. &
    Direct profile; 20 Hz states and 100 Hz control records. \\[2pt]
    Game configuration &
    PickLift instructions; five attempts, one target, one-turn cooldown. &
    Drawer instructions; five attempts, 30 damage per target, one-turn cooldown. \\[2pt]
    Shared bindings &
    \multicolumn{2}{p{0.77\linewidth}@{}}{Instruction display; reset/start; controller dispatch; outcome-to-effect mapping; episode finalization.} \\[2pt]
    Integration output &
    Loadable PickLift resources and task descriptor, linked to the configured skill. &
    Loadable drawer resources and task descriptor, linked to the configured skill. \\[2pt]
    Required checks &
    Reset and lift boundaries; state/action alignment. &
    Asset/reset integrity; predicate boundaries; Direct timing and export alignment. \\
    \hline
  \end{tabularx}
\end{table*}

\subsection{Task Integration: Inputs and Outputs}
\label{sec:integration}

Task contracts, task adapters, and game configurations connect robot tasks to
a shared gameplay and collection workflow.

\paragraph{Contracts, adapters, and game configurations.}
The task contract declares robot and scene assets, control and reset interfaces,
initial-state sources, success conditions, and the meaning and format of recorded
observations and controls. The task adapter implements or invokes the corresponding
controller, reset procedure, and success evaluation. A separate authored game
configuration associates the task with teaching content, skill parameters, and
course placement. Game settings can therefore change while retaining the task's
declared control and success semantics.

\paragraph{Shared workflow and outputs.}
Shared bindings coordinate preparation, execution dispatch, outcome delivery,
and episode finalization, calling adapters through the declared interfaces.
Game configurations select teaching content and map task outcomes to game effects.
Integration produces loadable task descriptors and resources bound to the
adapters and game settings. Existing batch tools generate supported descriptors,
parameter headers, resource packages, and provenance hashes. The same task
outcome drives game feedback and labels the episode; recorded observations and
controls retain the meanings specified by the task contract.
Section~\ref{sec:collection} details the recording and export format.

\paragraph{Adaptation and checks.}
Within supported interfaces, authors supply task resources, initial-state and
goal settings, teaching content, and game parameters; preparation, dispatch,
outcome delivery, and episode finalization are reused. Control interfaces,
success predicates, or recording formats not supported by existing adapters
require adapter extensions and review before integration. Checks cover assets,
resets, predicate boundaries, runtime behavior, and export alignment.

Table~\ref{tab:integration-examples} gives SO101 grasp-and-lift / LIBERO
middle-drawer bindings. The latter uses a versioned Direct profile rather
than LIBERO's operational space control (OSC) controller. The configuration
values are versioned examples; listed checks are obligations, not blanket
evidence that every task configuration has passed.

\paragraph{Task and recording coverage.}
The task catalog contains 130 LIBERO and four SO101 tasks, all with default
combat configurations. Pre-release collection yielded 2,141 unique episodes
across 106 LIBERO tasks; Section~\ref{sec:evaluation} evaluates policy learning
on the 66 LIBERO-90 tasks with VR data. These statistics describe configuration
coverage, recorded task coverage, and learning-evaluation coverage, respectively.

\subsection{Task Onboarding and Operation Guidance}
\label{sec:mastery}

\begin{figure*}[t]
  \centering
  \includegraphics[width=\linewidth]{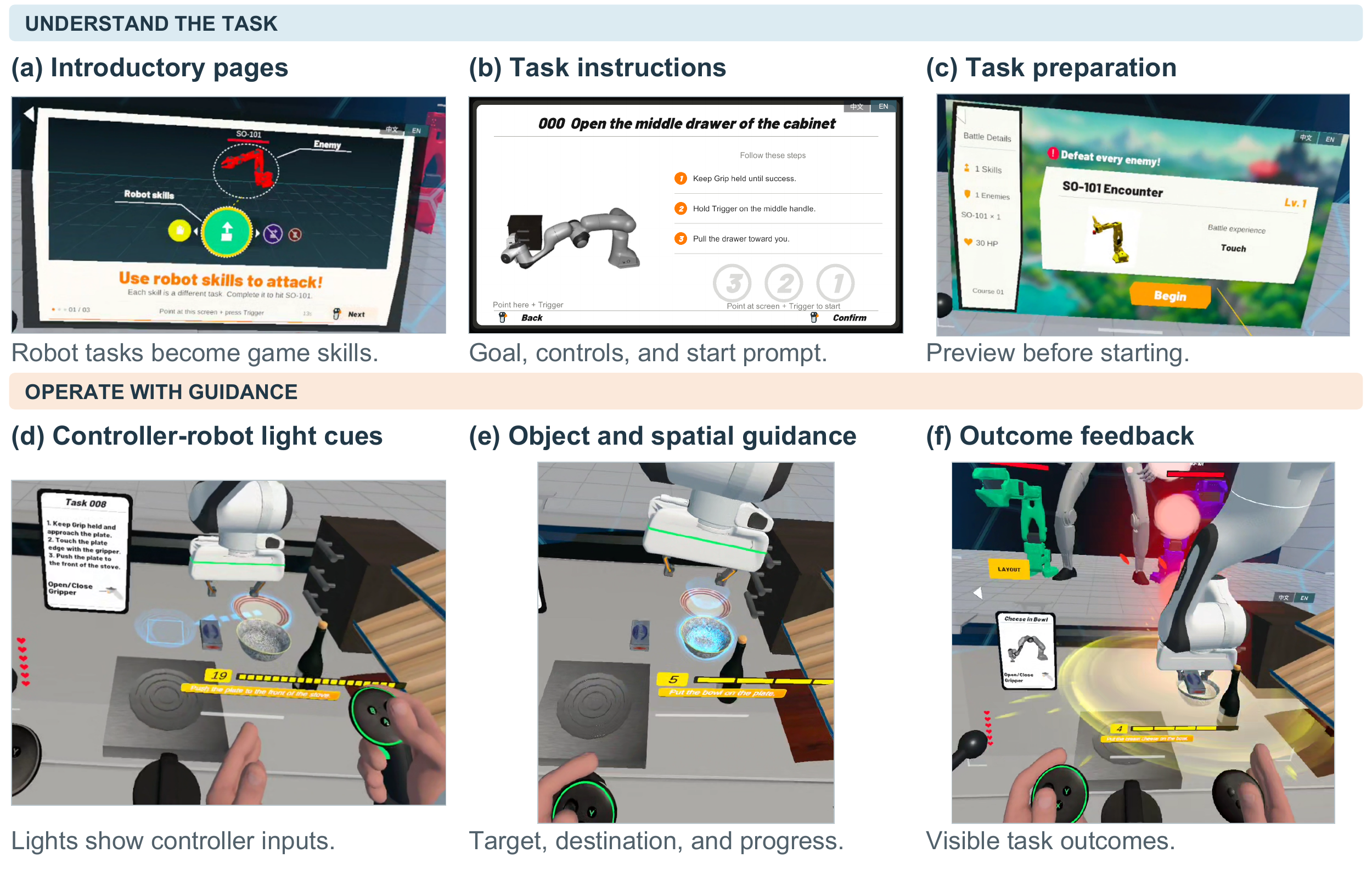}
  \caption{Task onboarding and operation guidance. (a) Introductory pages
  explain robot tasks as game skills. (b) Task instructions provide a goal,
  illustration, control steps, and start prompt. (c) A preparation screen
  previews the selected skill. During operation, (d) controller--robot light
  cues expose input states, (e) object highlights and spatial cues guide
  manipulation alongside the attempt count, and (f) visual effects provide
  outcome feedback. Panels show representative states from different tasks.}
  \label{fig:task-instruction}
\end{figure*}

RoboJackVR adds five complementary forms of assistance around imported LIBERO
tasks to support task understanding and robot operation.
\textbf{(1) Introductory pages} explain robot tasks as game skills, how to select
and start them, and how to follow demonstrations.
\textbf{(2) Controller--robot light cues} synchronize the virtual controller's
Grip and Trigger indicators with lights on the gripper housing and fingers,
making control engagement and grasping inputs visible on both sides of the
interaction.
\textbf{(3) Task instructions and demonstrations} present the goal and three
operation steps on a preparation screen, with looping example videos where
provided and illustrations otherwise (Figure~\ref{fig:task-instruction}).
\textbf{(4) Object highlights and spatial cues} use pulsing outlines to identify
the relevant object or handle, and directional arrows, destination regions,
or translucent reference poses to show how and where to move it.
\textbf{(5) Outcome feedback} communicates successful and failed attempts
through visual cues, sounds, and haptics, and retains the goal and guidance
for subsequent attempts when play continues. These aids share a presentation
flow configured with task-specific instructions, media, and scene bindings;
task completion remains governed by the declared success predicates.
For the middle-drawer task, these
steps instruct the player to keep Grip held, close the gripper around the
middle handle using Trigger, and pull the drawer toward them. The image
identifies the robot and object; the steps connect controller inputs to
the intended manipulation. Authors supply task-specific text and media,
while the instruction display and preparation flow are shared across tasks.

On that screen, players can read the
instructions, return to selection, or confirm their readiness. Once both
confirmation and task preparation are complete, a countdown leads into
the task, which waits for Grip input to begin control. Reading or leaving
the preparation screen does not start a recorded attempt. Tutorial viewing
progress is tracked separately from successful task execution.

\subsection{Game Loops and Progression for Continued Demonstration Collection}
\label{sec:progression}
\label{sec:gameplay}

The game combines tactical choices, battle pacing, and persistent progression
to organize repeated demonstration collection around immediate feedback and
subsequent goals. Robot tasks serve as selectable skills in a shared
adventure-game loop, inspired by the operation-to-skill structure of
\emph{Ring Fit Adventure}~\cite{ringfit2019}.

\paragraph{Robot operations as skill execution.}
Robot operations execute game skills, and damage, target count, cooldown,
and failure consequences make task selection depend on the battle state.
Players use VR controllers to grasp, place, stack, or open drawers. A
\emph{skill} is a robot-task challenge performed through teleoperation, not a
learned policy. One skill execution comprises an \emph{attempt group}: a
configured number of attempts at that task. Successful attempts trigger
the configured combat effect.

System faults use separate termination labels. An unfinished attempt
interrupted by a system fault may consume one attempt slot. Once an outcome
has been registered, retrying its save does not apply damage again or consume
an additional slot. Returning from task preparation to skill selection does
not advance cooldown.

\begin{samepage}
An attack assigns each target $k$ an integer budget $B_k\geq0$ over $N$
attempts. Integer shares $b_{j,k}\geq0$ sum to $B_k$ and differ by at most one.
For remaining health $H_{j,k}$ and success $y_j\in\{0,1\}$, attempt $j$ deals
\begin{equation}
  d_{j,k}=\min\!\left(H_{j,k},\,y_j b_{j,k}\right).
  \label{eq:task-damage}
\end{equation}
Each selected target has its own budget; it is not
split across targets. Ordinary failure consumes a slot with zero damage.
Unused shares after failure or an early exit are not redistributed.

\end{samepage}

For a configured cooldown of $c>0$ player turns, the first registered
attempt outcome sets an internal counter to $c+1$. Each complete enemy turn
then decrements it once; the skill becomes selectable when the counter reaches
zero. Thus, for $c=1$, using the skill sets the counter to 2; the immediately
following enemy turn reduces it to 1, so the next selection excludes that skill.
After a different skill or a rest turn and the next complete enemy turn, it
reaches 0 and the skill is available again. A zero-cooldown skill has no lockout.
If enemies remain after an attempt group, their turn precedes the next selection. These rules place repeated robot operations within combat
decisions: players choose targets, plan skill use, and face the consequences
of enemy attacks. The fault and save rules above prevent duplicate game
consequences during record recovery.

\paragraph{Task attempts within a battle.}
A task attempt is the basic operation unit: several attempts can execute a
selected skill, and skills and enemy turns form a battle. Attempt groups
alternate with enemy turns and separate defense interactions, placing repeated
robot operations within a sequence of choices, actions, and responses.

Players can choose among currently available skills or use free collection,
which provides single or grouped attempts and a return to task selection.
These entry and exit points let players choose both content and activity
length while retaining explicit boundaries for the recorded attempts.

\begin{figure*}[t]
  \centering
  \includegraphics[width=\linewidth]{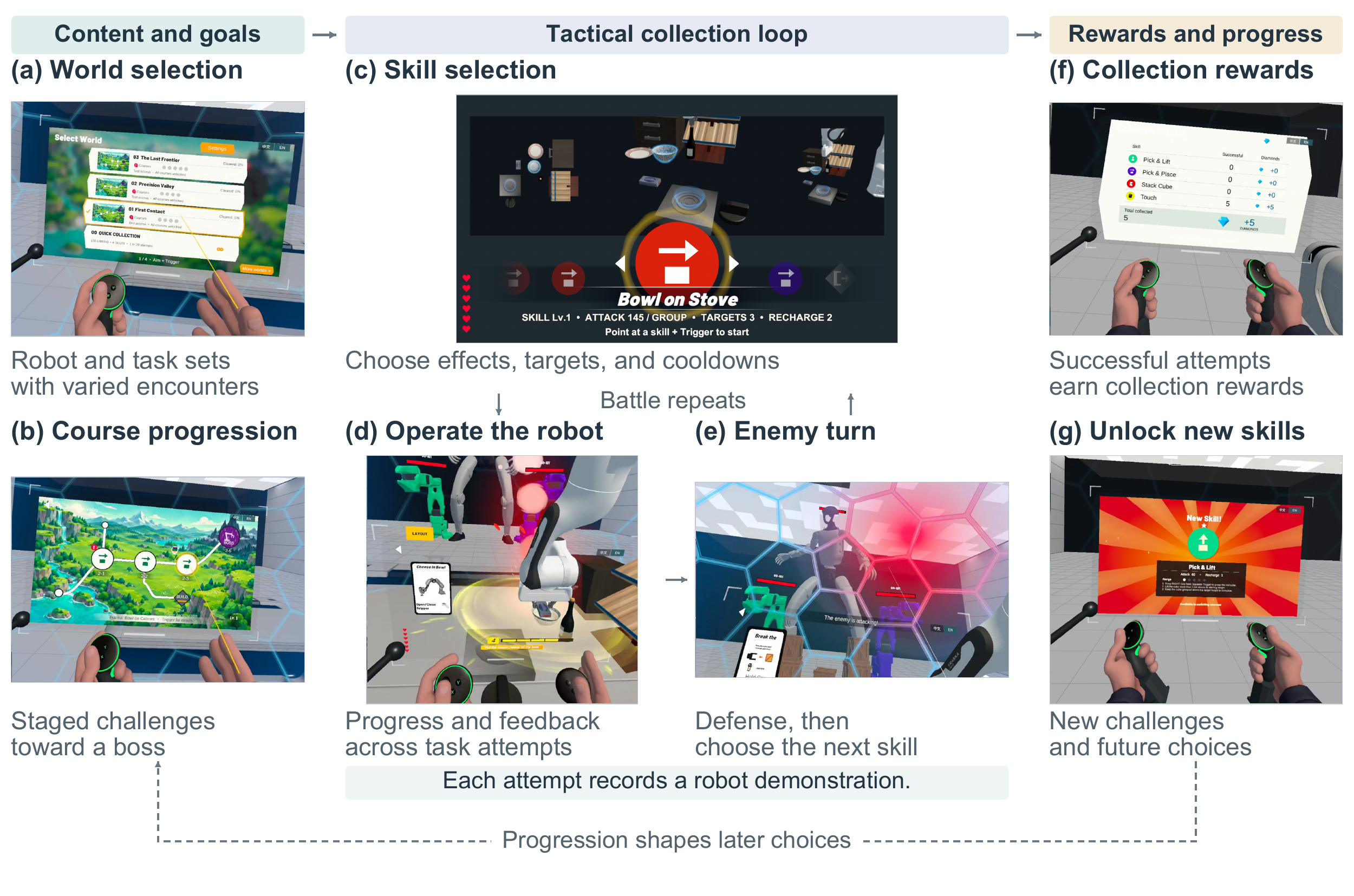}
  \caption{Game loops supporting demonstration collection. (a--b) Worlds
  and courses organize tasks and encounters. (c--e) Players select a skill,
  execute its robot-task attempts with progress and outcome feedback, and
  respond to the enemy turn before choosing again. (f--g) Collection rewards
  and skill unlocks connect contributions to later play. Task execution also
  produces demonstration records. Screens show different tasks and play stages.}
  \label{fig:gameplay-loop}
\end{figure*}

\paragraph{Progression and subsequent challenges.}
Battle outcomes update experience, levels, health capacity, and world/course
access, leading toward boss encounters. These changes carry the consequences
of one battle into the content and challenges available afterwards. Newly
integrated tasks add challenges while reusing instruction, feedback, combat,
progression, and recording. The current robot-arena narrative connects these
challenges through longer-term goals. Figure~\ref{fig:gameplay-loop} shows how worlds and courses \mbox{organize}
challenges, how skill choices and task attempts alternate with enemy turns,
and how collection rewards and skill unlocks connect to later play.

\paragraph{Game mechanics and demonstration distribution.}
During combat collection, game rules also shape which tasks players select
and repeat. Course unlocks affect which tasks are accessible, skill cooldowns
restrict choices on the current turn, and attempt grouping and battle
termination affect the number of records produced. For example, once all
enemies are defeated, the system saves the current attempt and ends the
battle; remaining attempts that have not started are not executed. Task
difficulty, execution time, and combat payoff may also influence skill
preferences. Task proportions and repetition counts in the collected data
therefore need to be interpreted in relation to game configuration, battle
state, and player choices; game parameters form part of the demonstration
collection setup.

\subsection{Demonstration Collection}
\label{sec:collection}

Recording follows robot-task boundaries and retains the control profile needed
to interpret each demonstration. Prepared attempts are saved as separate
episodes ending in success, failure, or interruption; preparation faults can
occur before any episode exists. Metadata associates records with operator,
session, task, model, and controller identities. The Franka schema also has
collection-entry and context fields; analyses comparing combat and free
collection must check their availability in the selected records.
Skill choice and defense stay outside the robot action sequence.

\paragraph{A concrete export profile.}
The native Franka Direct exporter preserves 20 Hz observations and the five
ordered 100 Hz control ticks executed between adjacent observations:
\begin{equation}
  \mathcal{D}_e =
  \bigl((o_t,U_t,o_{t+1})\bigr)_{t=0}^{T_e-1},
  \qquad U_t=(u_{t,1},\ldots,u_{t,5}).
  \label{eq:demonstration-export}
\end{equation}
Here $o_t$ is the observation at step $t$, $U_t$ is the ordered control
sequence executed between consecutive observations, and $T_e$ is the number
of observation intervals. This representation preserves temporal alignment
between observations and executed controls. Franka Direct uses its own control
representation, which cannot be interpreted directly as LIBERO OSC actions;
the field definitions are as follows.

Robot state contains seven arm angles in radians and two finger positions in
meters. Each control tick retains joint targets in radians, applied actuator
controls, state, and timing; the applied controls comprise seven arm torques
(Nm) and two finger position targets (m). Raw row $t+1$ stores the controls
from observation $t$ to $t+1$. Its eight-component end-of-row target/trigger
field is diagnostic and does not replace this interval sequence or represent
a LIBERO OSC action.

\paragraph{Rendering and player-side records.}
Images are rendered from recorded scene states at matching observation
indices, using front and wrist cameras, without stepping dynamics.
This provides state-aligned views, not evidence that executing the saved
commands reproduces the trajectory. Player sidecars link head/controller
poses and inputs by timestamps and input sequence IDs, supporting process
analysis and approximate player-view reconstruction; full game-display
replay additionally needs presentation-state records.

The exporter checks integrity, tick order, and state/control alignment.
Task success, export validity, and training admission remain separate.
A learning adapter must declare its action representation, camera processing,
and timing rather than silently collapse the control block. The converted learning interface and selected subset are described in
Section~\ref{sec:evaluation}, together with the downstream policy evaluation.

\section{Policy Learning from VR Demonstrations}
\label{sec:status}
\label{sec:evaluation}

We ask whether play-collected demonstrations can replace or supplement
standard demonstrations. Before public release, 21 participants collected
1,310 VR demonstrations covering 66 LIBERO-90 tasks~\cite{libero2023}; we
train ACT~\cite{zhao2023act} from random initialization and evaluate on these
66 tasks. Standard data are LIBERO's simulation teleoperation demonstrations.

\begin{table*}[t]
\centering\small
\caption{Success on the 66 LIBERO-90 tasks with VR data, mean of four seeds. Held-out:
initial states of 10 held-out standard demonstrations per task (660 cases);
official: 5 official initial states per task (330 cases). Matched updates: 210k
updates. Bold / underline: best standard-only / VR arm.}
\label{tab:act-results}\label{tab:act-design}
\begin{tabularx}{\linewidth}{@{}Yrrrr@{}}
\hline
Training data & Demos/task & Updates & Held-out & Official \\
\hline
Standard 100\% & 40 & 335k & \textbf{76.6\%} & \textbf{73.5\%} \\
VR only & $\sim$20 & 127k & \underline{69.6\%} & 68.2\% \\
VR + standard 25\% (joint) & $\sim$20\,+\,10 & 210k & 66.6\% & \underline{69.2\%} \\
Standard 25\% & 10 & 83k & 58.2\% & 54.8\% \\
\hline
VR only, matched updates & $\sim$20 & 210k & 65.3\% & 68.7\% \\
Standard 25\%, matched updates & 10 & 210k & 54.8\% & 55.8\% \\
\hline
\end{tabularx}
\end{table*}

\subsection{Conversion to Standard Demonstrations}
Franka Direct records are converted to 20 Hz OSC trajectories in robosuite
1.4.1 / MuJoCo 2.3.7. The converter restores the initial state and executes
end-effector and gripper references reconstructed from the recorded controls,
rendering both RGB views. References
are slowed by a factor of two, and the arm pauses while the gripper opens or
closes. Trajectories are kept even when the converted execution fails the
task. The 1,310 converted episodes give about 20 per task.

\paragraph{Gripper format.} The converter emits gripper tracking events
in $\{-1,0,+1\}$, with $0$ meaning hold. The converted channel starts at $-1$
(open) when the initial finger gap exceeds 0.04\,m and at $+1$ (closed)
otherwise; after a $+1$ event it stays $+1$ until a $-1$ event, and vice versa.
Images and the six arm dimensions are unchanged.

\subsection{Training and Held-Out Evaluation}
\label{sec:act-setup}
\paragraph{Data.} Standard demonstrations are LIBERO's official LIBERO-90
HDF5 files, with 40 training and 10 validation demonstrations per task; the
validation demonstrations are held out as the test set and never used for
training. ``$k$ per task'' takes the first $k$
training demonstrations. VR demonstrations are the 1,310 converted episodes
covering 66 LIBERO-90 tasks, drawn from 2,130 technically valid conversions
across 106 LIBERO tasks. Episodes whose converted execution fails the task are
retained. Standard training sets span all 90 LIBERO-90 tasks; all reported
results are on the 66 tasks with VR data.

\paragraph{Model and optimization.} We adapt LeRobot ACT~\cite{lerobot2024}
with a task-ID input over the 130-task index space, two $128\!\times\!128$ RGB
views (agent view and eye-in-hand), a 9-D state (seven joint angles and two
finger positions) and 100-action chunks of 7-D OSC pose deltas plus gripper.
All weights are randomly initialized; no pretrained visual backbone is loaded.
AdamW uses learning rate $10^{-4}$, weight decay $10^{-4}$, 500 linear warmup
updates, cosine decay to $10^{-5}$, gradient clipping at 1.0, BF16 autocast and
batch size 16. Each epoch uses every training frame once as a window start,
shuffled without replacement. Training runs for 10 epochs and the final
checkpoint is evaluated without selection. Every arm is trained with seeds 20261012, 20261011, 20261010 and 20261009; tables report the mean.
States and actions are z-score normalized with statistics from each arm's own
training set; the joint arm uses statistics of the combined set.

\paragraph{Held-out evaluation.} For each LIBERO-90 task, the 10 validation
demonstrations in the official HDF5 file form the test set; they are excluded
from every training set and from normalization statistics. Each policy starts
from the recorded initial state of each held-out demonstration (660 cases per
arm). None of the held-out initial states coincides with a training demonstration,
the 50 official initial states, or the 5 official states used in
Table~\ref{tab:act-results}: minimum state distances to those sets are
0.022--0.058, comparable to the 0.024 between held-out states themselves. VR
episodes lack full simulator states, so they were compared by first-frame
images only (minimum pixel difference 1.47, versus 1.55 between standard
training and held-out frames); exact overlap with VR layouts cannot be ruled
out.
Control runs at 20 Hz; each prediction executes ten actions before replanning,
after ten settling steps with the gripper open. The horizon is 900 steps, a
project protocol rather than the official default. Success is the
environment's \texttt{check\_success}; numerical exceptions
(\texttt{FloatingPointError}) are not resampled and count as failures.

\paragraph{Official initial states.} The second test runs the same final
checkpoints from 5 official LIBERO initial states per task (330 cases per arm)
with the same control settings.

\subsection{Results}
\label{sec:act-results-text}

\paragraph{VR versus standard demonstrations.} Without any standard
demonstrations, VR-only training reaches 69.6\% (held-out) and 68.2\%
(official), 91\% and 93\% of the full-standard arm, with about 20 VR episodes
per task versus 40 standard demonstrations.
It also exceeds the 25\% arm by 11.4 and 13.4 points.

\paragraph{Supplementing reduced standard data.} Adding VR to the 25\%
standard set raises success from 58.2\% to 66.6\% (held-out) and from 54.8\%
to 69.2\% (official), 87\% and 94\% of full-standard performance with 75\%
fewer standard demonstrations.

\paragraph{Matched updates.} Trained for the joint arm's 210k updates, the
25\% arm reaches only 54.8\% and 55.8\%, below both VR arms, indicating that extra updates alone do not explain the VR gain in this comparison. VR-only training with 210k updates gives
65.3\% and 68.7\%: held-out success decreases, while official-state
success increases slightly relative to its 127k run.

\paragraph{Ablation: no gripper conversion.} Training on the same 1,310 VR
episodes while keeping the converter's gripper events, without the level
conversion described above, reaches 9.7\% on official and 7.6\% on held-out initial states, compared with
66.7\% and 68.5\% with the conversion under the same seed (20261012). This
ablation has one seed.

\subsection{Analysis and Limitations}
\label{sec:act-analysis}
VR has about twice as many episodes per task as the 25\% arm.
Conversion slows references and pauses the arm during
gripper motion, so VR data also differ in execution timing. The size of the
joint arm's margin differs between the two tests. Each arm has four seeds, in simulation, with one
architecture.

\section{Conclusion and Future Work}
\label{sec:limitations}
RoboJackVR connects reusable task contracts and shared bindings to guidance,
combat, progression and demonstration recording in a standalone VR game.
It includes 130 LIBERO and four SO101 tasks. On LIBERO-90, VR-only training
reaches 91\% of full-standard success on covered tasks, and adding VR to 25\% of the
standard data outperforms an update-matched 25\% control; four seeds and
altered execution timing limit this evidence. This pre-release study focuses on the implemented system
and its collection-to-learning pipeline; public deployment will enable studies
of participant diversity, onboarding, sustained contribution, integration
effort and physical-robot transfer.

\IEEEtriggeratref{20}
\bibliographystyle{IEEEtran}
\bibliography{references}
\end{document}